\documentclass[letterpaper,journal]{IEEEtran}
\usepackage{amsmath,amsfonts}
\usepackage{algorithmic}
\usepackage{algorithm}
\usepackage{array}
\usepackage[caption=false,font=normalsize,labelfont=sf,textfont=sf]{subfig}
\usepackage{textcomp}
\usepackage{stfloats}
\usepackage{url}
\usepackage{verbatim}
\usepackage{graphicx}
\usepackage{cite}
\usepackage{tabularx}
\usepackage{colortbl}  
\usepackage[dvipsnames,table]{xcolor}
\usepackage{float}       

\definecolor{darkorange}{RGB}{255,165,0}
\definecolor{darkblue}{RGB}{70,130,180}
\definecolor{darkgreen}{RGB}{34,139,34}
\definecolor{gray}{gray}{0.8}

\definecolor{orange}{RGB}{255,165,0}  
\definecolor{blue}{RGB}{0,0,255}      
\definecolor{green}{RGB}{0,255,0}     
\definecolor{gray}{gray}{0.8}

\begin{document}

\title{
Vehicle Localization in GPS-Denied Scenarios Using Arc-Length-Based Map Matching
}

\author{Nur Uddin Javed$^{1}$ and Yuvraj Singh$^{2}$ and Qadeer Ahmed$^{3}$
\thanks{*This work was not supported by any organization}
\thanks{$^{1}$Nur Uddin Javed is with the Department of Electrical and Computer Engineering,The Ohio State University,
        Columbus, OH 43210, USA
        {\tt\small javed.31@osu.edu}}%
\thanks{$^{2}$Yuvraj Singh is with the Department of Mechanical and Aerospace Engineering, The Ohio State University,
        Columbus, OH 43210, USA
        {\tt\small singh.1250@osu.edu}}%
\thanks{$^{3}$Qadeer Ahmed is with the Department of Mechanical and Aerospace Engineering, The Ohio State University,
        Columbus, OH 43210, USA
        {\tt\small ahmed.358@osu.edu}}%
\thanks{$^{1,2,3}$The authors are affiliated with the Center for Automotive Research, The Ohio State University,
        Columbus, OH 43212, USA}%
\thanks{This work has been submitted to the IEEE for possible publication. Copyright may be transferred without notice, after which this version may no longer be accessible.}
}

\markboth{Journal of \LaTeX\ Class Files,~Vol.~14, No.~8, August~2021}%
{Shell \MakeLowercase{\textit{et al.}}: A Sample Article Using IEEEtran.cls for IEEE Journals}


\maketitle

\begin{abstract}
 Automated driving systems face challenges in GPS-denied situations. To address this issue, kinematic dead reckoning is implemented using measurements from the steering angle, steering rate, yaw rate, and wheel speed sensors onboard the vehicle. However, dead reckoning methods suffer from drift. This paper provides an arc-length-based map matching method that uses a digital 2D map of the scenario in order to correct drift in the dead reckoning estimate. The kinematic model's prediction is used to introduce a temporal notion to the spatial information available in the map data. Results show reliable improvement in  drift for all GPS-denied scenarios tested in this study. 
 This innovative approach ensures that automated vehicles can maintain continuous and reliable navigation, significantly enhancing their safety and operational reliability in environments where GPS signals are compromised or unavailable.
\end{abstract}


\section{Introduction}

\IEEEPARstart{L} {ocalization} is a safety-critical requirement for automated driving system. GPS is widely used as reliable of information \cite{vanet_survey,zhi,faroog,bajaj02,alami}. However, in GPS-denied environments such as signal disruptions in urban areas, alternative methods like dead reckoning and IMU can be used to provide a localization estimate. However, dead reckoning and IMU estimates suffer from drift that can only be corrected using fusion with GPS.

This paper describes a method to fuse the kinematic model's temporal information with the spatial 2D map of the driving environment thus enabling localization in GPS-denied scenarios without using relying on communication networks like 5G, V2X, WiFi, Zigbee, etc.

\subsection{Literature review}

For automated driving applications, vehicles have been equipped with diverse combinations of on-board vehicle sensors, GPS and off-board information from communication techniques like V2X, 5G network \cite{big_survey}. Each hardware configuration comes with its own set of strengths and limitations owing to real-life scenarios like intersections, environmental conditions, traffic density, cost, etc.
Vehicle localization methods can be classified into four types based on their level of reliance on GPS availability \cite{vanet_survey}:
\begin{enumerate}
    \item \textit{GPS-only methods:} uses satellite signals to determine vehicle location but signal obstruction often causes delays, resulting in unreliable localization.\cite{vanet_survey,bajaj02}.
    
    \item \textit{GPS-Based Cooperative Localization:} combines GPS with other information sources like Inertial Measurement Units (IMU), Camera, Light Detection and Ranging (LiDAR), Vehicle-To-Everything (V2X) communication, etc., and use techniques like map matching, and metrics like, Time of Arrival (TOA), Arrival of Angle (AOA)  are employed to enhance GPS position accuracy. \cite{zhi,vanet_survey}. 
        Notably, vehicle-to-vehicle (V2V) communication using TOA metrics enhances GPS Accuracy\cite{alami}. 
        Utilizing  Lane Detection vision-based localization achieves  centimeter-level precision with a mean error of 0.73 m to enhance GPS accuracy through  its performance dependent on 3D map, availability lighting and viewing angles \cite{Kamijo}. 
        In parallel, Rader based SLAM  has been developed, providing high accuracy with an RMSE 0.07 m lat and 0.38 m long to enhance the accuracy of GPS through a lack of resilience in dynamic environment \cite{Ward}. Additionally,Cooperative map matching significantly  reduces  GPS error, decreasing from 15.48 m to 6.31 m \cite{Rohani}. Collectively, this methods represent significant strides toward overcoming GPS limitation.
    
    \item \textit{Non-GPS-Based Cooperative Localization:} 
        Cooperative methods operate independently from GPS \cite{vanet_survey}. 
        Wahab et al. \cite{wahab} demonstrated that the Time of Arrival (TOA) metric applied to signals from Roadside units (RSUs), can effectively determine vehicle positioning.
        However, this approach is influenced by factors such as vehicle's speed and traffic density, which increases error of the localization estimate.
        Lidar-based SLAM has  shown high accuracy achieving  0.017 m lat and 0.033 m long RMSE although it depends on the availability of 3d prior map \cite{Castorena}
        Similarly, Lidar map based visual localization demonstrated high precision with errors  0.014 m lat and 0.019 m long  \cite{Wolcott}.

        \IEEEpubidadjcol
           
    \item \textit{Non-Cooperative Localization:} 
        These methods rely on a single source of information.
        GPS-independent network-based vehicle localization systems using directional antenna-equipped RSUs along roads have been employed where signals from RSUs are used by vehicles to localize itself using only RSU beacon messages \cite{ou}. However,this approach is not widely applicable in the real world due to its reliance on standard antenna patterns and has performance issue at high speed.
        Zigbee and TOA-based positioning approaches, including C-V2X \cite{wang, sameh}, offer alternatives, through those face reliability issues in GPS-denied areas.
        Sensor-based non-cooperative localization, such as using LiDAR for lane marking and localization, offers precision but requires real-time data processing \cite{ahmed,ohta}.
        Dead reckoning another non-cooperative localization approach offers relatively precise positioning and navigation for vehicles in the absence of other sensors .However, it is prone to error accumulation or drift when GPS is unavailable \cite{welte,freydin}.
        Carlson et. al. \cite{carlson} analyzed  the error factors in land vehicle dead reckoning system that uses differential wheel speeds  unaccounted  roll dynamics and driving surface variations as major sources of error. The proposed method performed well during GPS outages by addressing the limitation. 
        Welte et al. \cite{welte} proposed a method to accurately calibrate system parameters without prior knowledge of the actual conditions using Rauch-Tung-Striebel smoothing to  keeps drift under 1 meter over a distance of 100 meters, thereby enhancing the precision of dead reckoning systems.
        Similarly, Freydin et. al. \cite{freydin} developed a deep neural network (DNN) that estimates vehicle speed from inertial sensor data  reducing position error to less than 120 m during a 4 minute driving mission.
        In another approach, Brossard et al. \cite{brossard} introduced a DNN-based inertial dead reckoning method  with an adaptive Kalman filter to improved short-distance accuracy.
        Kaise's patent \cite{kaise} discusses a cooperative technique that combines radio and self-reckoning methods for reliable positioning.
        Furthermore, Cao \cite{cao}  examined the limitations of dead reckoning in high-speed scenarios experimentally, particularly in GPS-denied environments such as  tunnels. Dead Reckoning based on vehicle kinematics was found to suffer from severe severe limitations at high speeds. This indicated additional method or correction is to improve performance in such GPS denied environment

\end{enumerate}

\subsection{Problem Definition and Novel Contributions}

\begin{figure}[ht]
    \centering
    \includegraphics[width=0.8\linewidth]{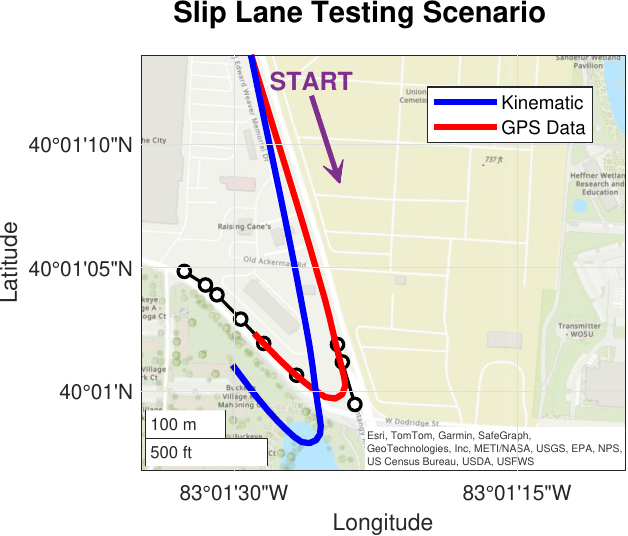}
    \caption{Dead reckoning error accumulation: Slip-lane scenario}
    \label{fig:Dead_Reckoning_Error_Accumulation}
\end{figure}

In GPS-denied scenarios, network-based cooperative methods may not be widely available everywhere \cite{ou,wang,sameh}. 
Camera and LIDAR based solution has its own limitations.Hence, dead reckoning using onboard sensors becomes the primary localization method in most situations in practice. However, dead reckoning estimates accumulate positioning error over time as compared to the ground-truth localization as seen in Figure \ref{fig:Dead_Reckoning_Error_Accumulation}, where the test vehicle is driven on a road that has a slip-lane intersection.
As observed, the drift is most severe in the lateral direction of the vehicle. In traditional localization frameworks, filter methods such as Kalman Filters (KF), Extended Kalman Filters (EKF), and Bayesian filters are used to estimate a vehicle's state by recursively integrating dynamic measurement inputs to correct predicated states derived from dead reckoning. These filters operate through a prediction update cycle where the state prediction $\hat{x}_{k}^-$ obtained from a discrete-tme process model of the system is refined through measurement  $z_k$ via equations as shown below: 
\begin{equation}
    \hat{x}_k = \hat{x}_k^- + K_k (z_k - H_k \hat{x}_k^-),
\end{equation}
with $K_k$  representing Kalman gain \cite{Kalman1960, Simon2006}. The \textit{Extended Kalman Filter (EKF)} extends this to handle non-linear state and measurement models by linearizing them around the current estimate \cite{Julier1997}.
Bayesian filters encompass a broader class of methods including particle filters which provide a probabilistic framework for state estimation without strict linearity or gaussian noise assumptions \cite{Ristic2004}. In GPS-denied environments, dead reckoning using vehicle's onboard sensor data available on the Controller Area Network (CAN) bus is only source to provide high-frequency sensor data required for state estimation. But, dead reckoning estimation using CAN data is inherently prone to cumulative drift as described by the state transition model:

\begin{equation}
    x_k = F_k x_{k-1} + B_k u_k + w_k \quad \text{\cite{Simon2006}},
\end{equation}

This drift leads to an unbounded growth in the error covariance $P_k$:
\begin{equation}
    P_k = F_k P_{k-1} F_k^T + Q_k,
\end{equation}
This unbounded growth in $P_k$ diminish the filters ability to maintain accurate localization over time. While EKF can handle certain nonlinearities and Bayesian Filters offer more flexibility in noise modeling, both rely on dynamic, temporally varying measurement to correct state estimates. In the absence of GPS, IMU and static map data are available to use as measurement. However, IMU-based localization is prone to drift if sensor fusion with GPS is not performed. These filters cannot effectively mitigate the drift of dead reckoning estimate that uses vehicle sensor data, hence leading to inaccurate localization. Moreover,attempting to use map data as measurement introduce additional challenges, as maps data are static and lack the temporal variability required for meaningful state corrections within filtering frameworks. Consequently without reliable dynamic measurements to correct state, Kalman Filters, EKF, and Bayesian Filters are unable to sustain accurate localization in GPS denied scenarios.
This paper addresses these challenges by proposing a arch length based  map matching technique to correct the  drift and improve localization accuracy in GPS-denied scenarios.

\subsection{Testing Scenarios}
For algorithm verification and validation, six test maneuvers were done on various road geometries, as outlined in table \ref{tbl:test_scenarios} under GPS denied. Scenarios 1 and 2 both occurring at a  four-way intersection, involve right-turn  and left-turn maneuvers to verify the performance of  90 degree turn at low speed. In contrast, scenario 3 at the four way in same intersection evaluates performance at 45 mph (20.1 m/s) high speed run. Additionally, Scenario 4, set at a Y-intersection is designed to test mild and sharp turn geometries at low speed for right turn maneuver. Scenario 5 focuses on a roundabout, assessing the algorithm's handling of  constant curvature at low speed. Lastly, the slip lane tests vehicle behavior during high-speed entry into the lane followed by a low-speed cross maneuver.

\begin{table*}[ht]
\caption{Test Scenarios and Maneuvers}
\label{tbl:test_scenarios}
\renewcommand{\arraystretch}{1.4} 
\setlength{\extrarowheight}{1pt} 
\arrayrulecolor{black!30} 
\rowcolors{1}{gray!10}{white} 
\begin{tabularx}{\linewidth}{p{0.08\textwidth} p{0.20\textwidth} p{0.35\textwidth} >{\centering\arraybackslash}X}
\hline
\rowcolor{gray!20} 
\textbf{Scenario ID} & \textbf{Scenario Description}         & \textbf{Test Objective: Vehicle Dynamic Behavior Tested}                 & \textbf{Maneuvers}                                                  \\ \hline
1                    & Four-way intersection                & Low-speed 90-degree turn geometry                                        & \textbf{Right turn}                                                  \\ \hline
2                    & Four-way intersection                & Low-speed 90-degree turn geometry                                        & \textbf{Left turn}                                                   \\ \hline
3                    & Four-way intersection                & High-speed run (45 mph, 20.1 m/s)                                        & \textbf{Straight run}                                                \\ \hline
4                    & Y-intersection                       & Low-speed, mild, and sharp turn geometry                                 & \textbf{Right turn (sharp turn)}                                     \\ \hline
5                    & Roundabout (traffic circle)           & Road with constant curvature at low speed                                & \textbf{Straight}                                                    \\ \hline
6                    & Slip lane                             & Variable curvature turn without stopping                                 & \textbf{Right turn}                                                  \\ \hline
\end{tabularx}
\renewcommand{\arraystretch}{1} 
\end{table*}

A 2D map of the road geometry in latitude and longitude coordinates is typically available in the form of digital maps (OpenStreet Map, Google Maps). 
Hence, to correct the dead reckoning localization estimate, a novel arc length computation based map matching technique is proposed which matches the localization estimate with the geometry of the lane on which the vehicle is traveling.
A preliminary validation of the approach  was conducted by comparing it against a highly precise NovAtel PwrPak7D GPS/GNSS receiver \cite{novatel} having an accuracy of 2.5 cm throughout the test performed on three different road geometries.

\subsubsection{Assumptions}
The lane on which the vehicle is traveling is assumed to be known since automated vehicles are generally equipped with a perception system for detecting lanes.
Moreover, the vehicle is assumed to travel along the lane center line. which is reasonable for vehicle equipped with level 2 or higher automation capability.
The road geometry of the road/lane is treated by connecting various points on the lane in a piecewise linear manner, thus ensuring an efficient data structure. This limitation arises from the current assumption that the vehicle remains within a single lane, thereby simplifying the localization process but restricting the method's applicability in scenarios involving lane transitions.

\section{Kinematic dead reckoning: map matching based correction approach} \label{sec:dead_reckoning}
Consider a test vehicle of wheelbase $l$ located at coordinates $(x(t),y(t)) \in \mathbb{R}^2$ and oriented at a yaw angle $\psi(t) \in S^1 := [0,2\pi)$ at time $t$ with respect to an arbitrary inertial frame of reference. The vehicle's front wheels are rotated at steering angle of $\theta_f(t) \in S^1$ with respect to the vehicle's body. 
The generalized coordinates $q(t) = \{x(t),y(t),\psi(t),\theta_f(t)\}$ are the minimum set of independent coordinates that define the configuration of the vehicle system \cite{modern_robotics}. Generalized velocities $\dot{q}(t)$ are defined as the time derivatives of the generalized coordinates of the system. 
A kinematic model of the vehicle, $\dot{q}(t) = J(q(t))u(t)$, where $J(q(t))$ is a matrix-valued function of $q(t)$, is shown in equation \eqref{eq:kinematics} can be constructed by assuming rolling without slipping constraint at the wheel \cite{modern_robotics}.
\begin{equation}
    \begin{pmatrix}
        \dot{x}(t) \\ \dot{y}(t) \\ \dot{\psi}(t) \\ \dot{\theta_f}(t) 
    \end{pmatrix}
    = 
    \begin{pmatrix}
        \cos{\psi(t)}  &  0   \\
        \sin{\psi(t)}  &  0  \\
        \dfrac{\tan{\theta_f(t)}}{ l } & 0  \\
        0 & 1 
    \end{pmatrix}
    \begin{pmatrix}
        v(t) \\ \omega_f(t) 
    \end{pmatrix}
    \label{eq:kinematics}
\end{equation}

where, $v(t)$ is the velocity vector at the vehicle's rear axle, and $\omega_f(t)$ is the steering rate at the front wheels.

Onboard vehicle sensor data logged from the vehicle's Controller Area Network (CAN) bus (refer section \ref{sec:sensors}) provides the inputs $u(t)$ to compute the generalized velocities $\dot{q}(t)$ using the kinematic model $\dot{q}(t) = J(q(t))u(t)$ (equation \eqref{eq:kinematics}). Additionally, sensor measurements are used to update the matrix $J(q)$ at every time step. A dead reckoning estimate at time $t$ is obtained when the vehicle's current configuration $q(t)$ is computed using numerical integration as described in equations \eqref{eq:DR_location} and \eqref{eq:DR_yaw_angle}.
\begin{align}
    \text{For flat space $\mathbb{R}^2$: } 
        &\begin{pmatrix} x(t) \\ y(t) \end{pmatrix} 
            = \int_0^t \begin{pmatrix}\dot{x}(t) \\ \dot{y}(t) \end{pmatrix} dt \label{eq:DR_location} \\
    \text{For $\psi(t) \in S^1$ space: } 
        & \psi(t) = \int_0^t \dot{\psi}(t) dt \pmod{2\pi}
    \label{eq:DR_yaw_angle}
\end{align}

\subsubsection{Instrumentation and data logging} \label{sec:sensors}
The kinematic model is developed for the test vehicle having a wheelbase of 2.675m. The following quantities are logged from the CAN bus: 
The steering wheel rate [deg/s] is divided by vehicle steering gear train's steering ratio to obtain the front wheel's steering rate $\omega_f(t)$. 
Vehicle speed $v(t)$ [km/h] at rear axle is computed from the wheel speed sensor measurement by using a nominal wheel radius. Steering angle $\theta_f(t)$ is obtained from the hand steering wheel angle [deg] measurement by dividing it by the vehicle's steering gear train ratio. 
Yaw rate [deg/s] is numerically integrated to obtain a measured estimate of yaw angle $\psi(t)$. 
These values of $\theta_f(t)$ and $\psi(t)$ to update the kinematic model (equation \eqref{eq:kinematics}) at every time step.

\subsection{Trajectory arc length based map matching}
\begin{figure*}
    \centering
    \includegraphics[width = 0.8\linewidth]{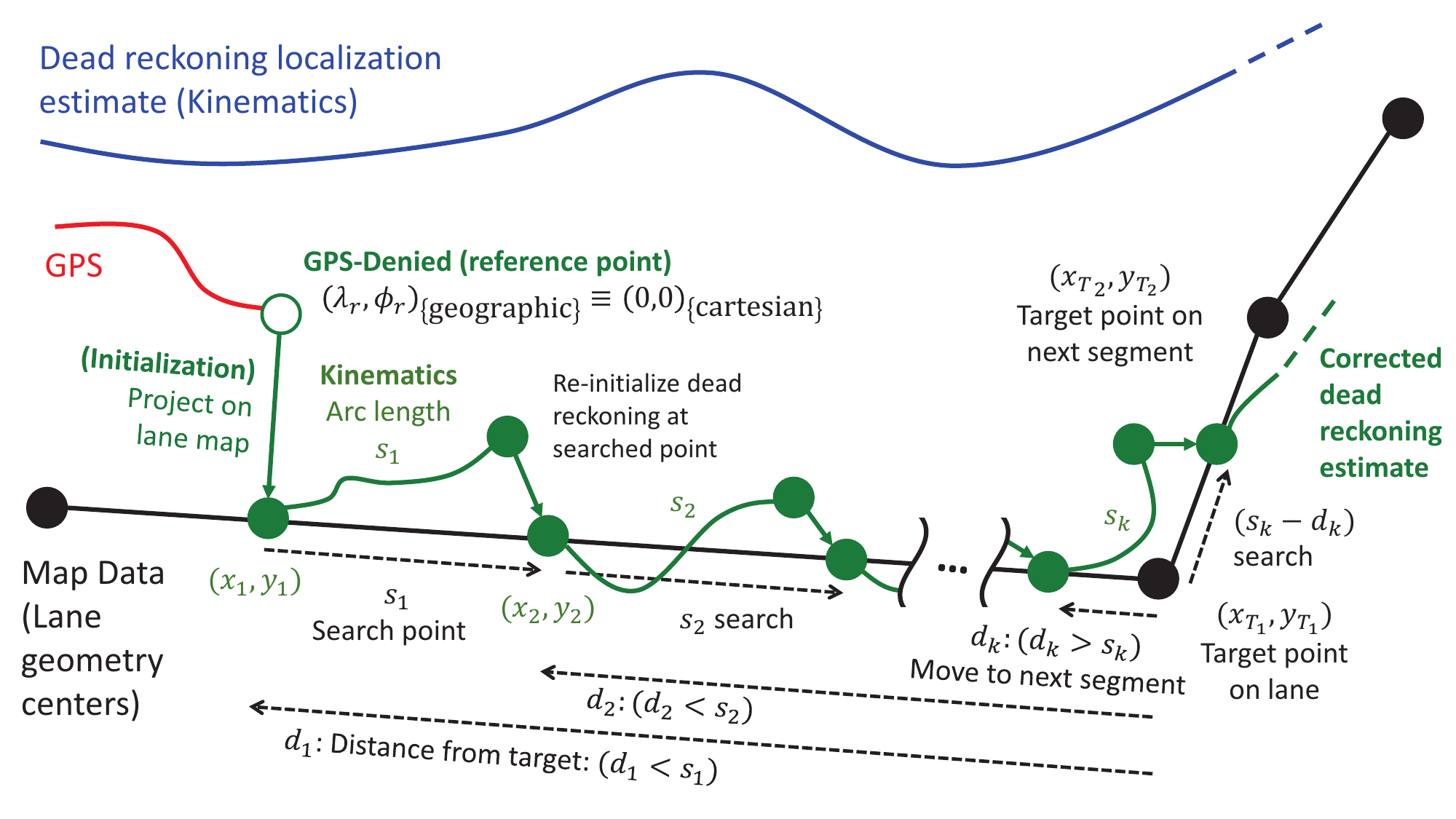}
    \caption{Correcting dead reckoning localization estimate through static map matching using arc length matching}
    \label{fig:map_matching}
\end{figure*}

The proposed approach considers static map information that consists of lane center coordinates of the road scenario. In a GPS-denied scenario, the dead reckoning localization estimate is corrected by map matching using lane information. 
Figure \ref{fig:map_matching} shows a schematic diagram describing the map matching algorithm's implementation. 

\begin{figure}
    \centering
    \includegraphics[width = \linewidth]{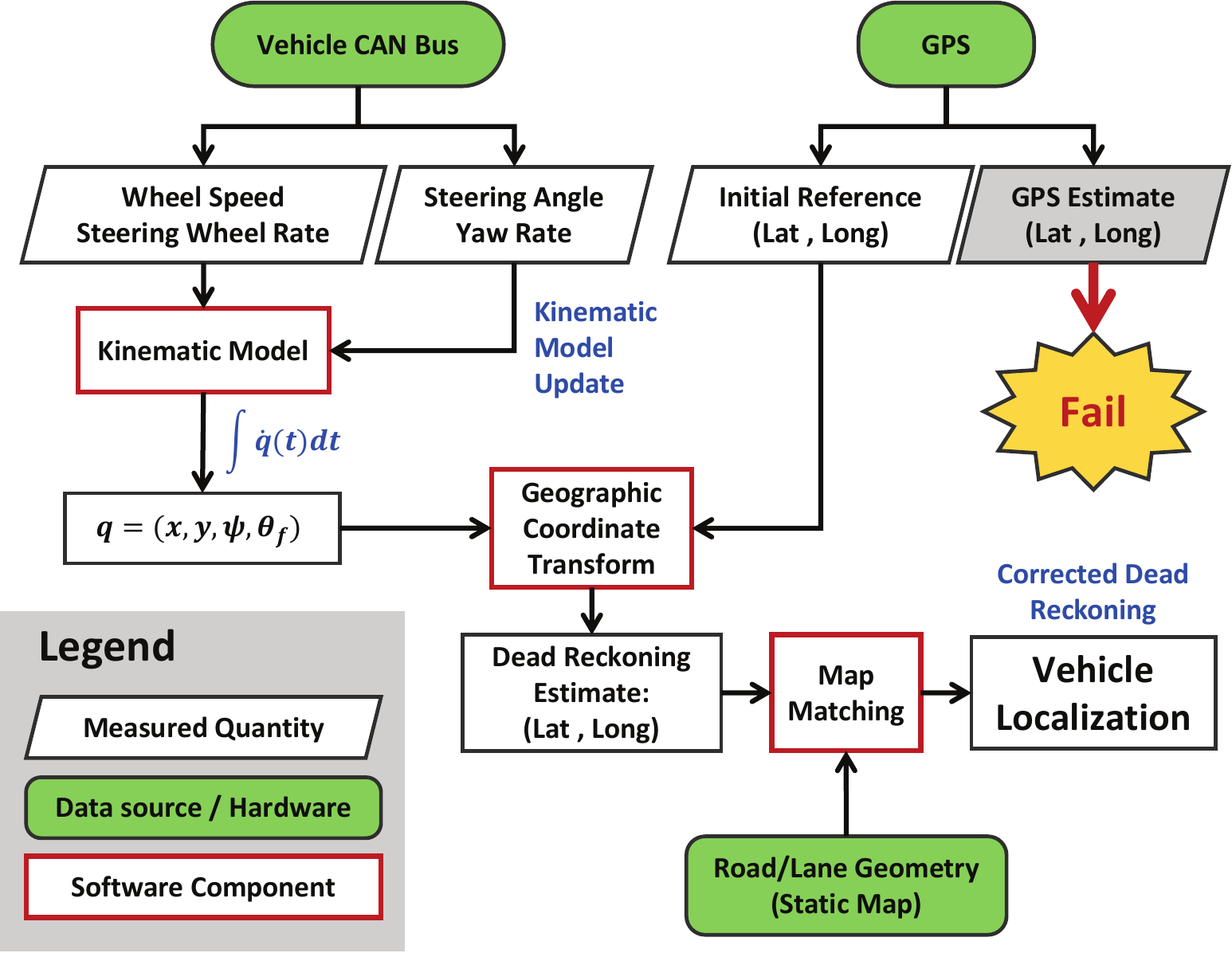}
    \caption{Overview of the proposed localization approach}
    \label{fig:Block_Diagram}
\end{figure}

\textbf{Notations:} The geographical coordinates (latitude, 
longitude) are denoted by $(\lambda,\phi)$, while the cartesian coordinates are denoted by $(x,y)$. The conversion between geographic coordinates and cartesian coordinates using equirectangular projection \cite{geocoordinates_book}.

\subsubsection{Initialization}
In the GPS-denied scenario, although the GPS estimate is unavailable beyond a certain point, there is a history of GPS points available. The last available location from GPS denoted by $(\lambda_r,\phi_r)$, , where $\lambda_r$ is the latitude and $\phi_r$ is the longitude of the reference point, is used as the initial reference localization estimate for the proposed algorithm. The slope of a line connecting the last two known GPS points is used to estimate the vehicle's initial yaw angle $\psi(0) = \tan^{-1}({\Delta y/\Delta x})$. The static map points represented in geographic coordinates are transformed to their corresponding Cartesian coordinate representation on the tangent plane to the Earth with origin fixed at the geographic coordinates $(\lambda_r,\phi_r)$.

Coordinates $(x_1,y_1)$ of the initial corrected dead reckoning estimate are obtained by projecting the point $(0,0) \equiv (\lambda_r,\phi_r)_{\text{\{geographic\}}}$ onto the line segment $\mathcal{L}_{T_0,T_1}$ (equation \eqref{eq:line}). The line segment connecting the static map point $(x_{T_{i}},y_{T_{i}})$ that the vehicle is targeting, and the static map point $(x_{T_{i-1}},y_{T_{i-1}})$ preceding it is shown in equation \eqref{eq:line}.
\begin{equation}
    \mathcal{L}_{T_{i-1},T_{i}} :=
    \begin{cases}
        \left\{ (x,y) \ \vert \ y = y_{T_i} + \dfrac{y_{T_i} - y_{T_{i-1}}}{x_{T_{i}} - x_{T_{i-1}}} \left( x - x_{T_{i}} \right) \right\} \\
        \left\{ (x_{T_i},y) \ \vert \ y \in [y_{T_{i-1}},y_{T_i}] \right\} \text{, for } x_{T_{i}} = x_{T_{i-1}}
    \end{cases}
    \label{eq:line}
\end{equation}

The coordinates $(x_1,y_1)$  are obtained using the orthogonality rule as shown in equation \eqref{eq:projection}.
\begin{equation}
    \left\langle \begin{pmatrix}x_1 \\ y_1 \end{pmatrix} , 
                 \begin{pmatrix} x_{T_1} - x_{T_0} \\ x_{T_1} - x_{T_0} \end{pmatrix}
    \right\rangle
    = 0 \text{ s.t. } (x_1,y_1) \in \mathcal{L}_{T_{0},T_{1}}
    \label{eq:projection}
\end{equation}

\subsubsection{Iteration and Re-Initialization of Kinematics}
The kinematic dead reckoning estimate is now computed for a batch of $N$ points using the equations \eqref{eq:kinematics}, \eqref{eq:DR_location} and \eqref{eq:DR_yaw_angle} for a batch of input velocity $v$ and steering rates $\omega_f,\omega_r$ (equation \eqref{eq:kinematics}) logged from CAN bus. 

For the $i$-th iteration, the point $(x_i,y_i)$ is considered as the initial point for the kinematics.
The arc length $s_i$ of the kinematic trajectory $(x(t),y(t))$ for the batch is computed from the generalized velocity $\dot{q}(t)$ using the Riemann integral approach \cite{spivak} as shown in equation \eqref{eq:arclength}, where $\lVert\cdot\rVert$ denotes the $\ell_2$ norm.
\begin{align}
    s_i &= \int_0^t{ \left\lVert \dfrac{d}{dt} \begin{pmatrix} x(t) \\ y(t) \end{pmatrix} \right\rVert dt }
    = 
    \lim_{\Delta t_k \rightarrow 0} \sum_{k=1}^{N-1} \left \lVert\begin{bmatrix} \dot{x}[t_k^*] \\ \dot{y}[t_k^*] \end{bmatrix} \right \rVert \Delta t_k  \label{eq:arclength} \\ 
    &\text{where, }
    t_k^* \in [t_{k},t_{k+1}] \ , \ \Delta t_{k}=t_{k+1}-t_{k} \nonumber
\end{align}
In equation \eqref{eq:arclength}, the time $t_k^*$ can be chosen arbitrarily within the bounds $[t_{k},t_{k+1}]$ since the generalized velocity $\dot{q}(t)$ computed from the kinematics is a continuous curve in time, thus is always Riemann integrable \cite{spivak}.


Now, the point $(x_{i+1},y_{i+1})$ is obtained by searching the point on the line segment $\mathcal{L}_{T_{i},T_{i+1}}$ having arc length $s_i$. The solution candidates $(x_{c_j},y_{c_j})\in \mathfrak{S}$ for search process are the roots to the quadratic polynomial $(x-x_i)^2+(y-y_i)^2 -s_i^2$ (refer equation \eqref{eq:arclength_search}), which implies that there are two roots that is, $\dim{\mathfrak{S}} = 2$. The solution $(x_{i+1},y_{i+1})$ accepted is the one that lies between the points $(x_i,y_i)$ and $(x_{i+1},y_{i+1})$. As a result, the solution with the shorter distance from the target point $(x_{T_i},y_{T_i})$ is selected as shown in equation \eqref{eq:choose_next_point_quadratic}.
\begin{equation}
    \mathfrak{S} \in \left\{(x,y) \in \mathcal{L}_{T_{i-1},T_{i}}  \ \vert \ 
        \left\lVert \begin{pmatrix} x-x_i \\ y-y_i \end{pmatrix} \right\rVert = s_i \right\}
    \label{eq:arclength_search}
\end{equation}
\begin{equation}
    (x_{i+1},y_{i+1}) = \mathrm{argmin}_{(x_{c_j},y_{c_j})\in \mathfrak{S}} \left\lVert \begin{pmatrix} x_{T_i} - x_{c_j} \\ y_{T_i} - y_{c_j} \end{pmatrix} \right\rVert
    \label{eq:choose_next_point_quadratic}
\end{equation}
The integrator for kinematics shown in equation \eqref{eq:DR_location} are re-initialized at $(x_{i+1},y_{i+1})$ for the next iteration.

\subsubsection{Moving to the Next static map Segment}

The distance between the current point and the target static map point is denoted by $d_i$ 
\begin{equation}
    d_i = \left\lVert \begin{bmatrix} x_i & y_i \end{bmatrix}^\top 
                    - \begin{bmatrix} x_{T_i} & y_{T_i} \end{bmatrix}^\top \right\rVert
\end{equation}
As the end of the line segment $\mathcal{L}_{T_{i-1},T_{i}}$ is reached, the distance $d_i$ will become shorter than the length $s_i$. When $s_i>d_i$, the algorithm moves to the next static map segment. The leftover length $(s_i - d_i)$ is searched on the next line segment $\mathcal{L}_{T_{i},T_{i+1}}$, and the algorithm's iterations continue as long as the static map data is available.


\begin{figure*}[t]
    \centering
    \includegraphics[width=\linewidth]{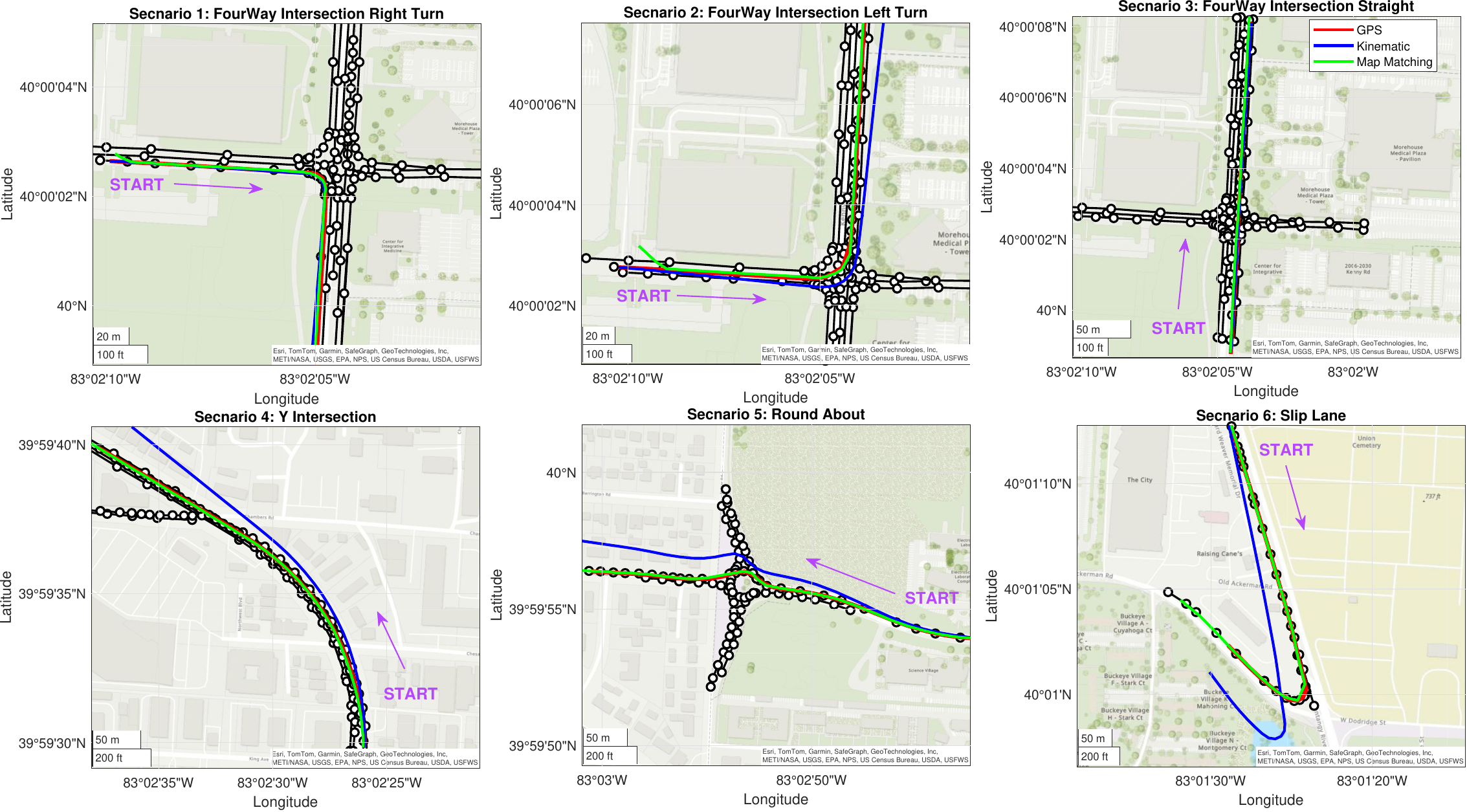} 
    \caption{Map Matching: Demonstration of spatial correction of the localization estimate for various traffic scenarios including intersections, roundabouts, and slip lanes. The map matching results are compared against GPS and dead reckoning paths to illustrate accuracy improvement.}
    \label{fig:Intersection_with_Map_Matching}
\end{figure*}

\section{Results and Discussion} 
\begin{table*}[htbp]
\caption{RMSE Comparison for Test Scenarios (X and Y Directions)}
\label{tbl:rmse_comparison}
\renewcommand{\arraystretch}{1.6}  
\setlength{\arrayrulewidth}{0.4pt}  
\arrayrulecolor{black!30}  
\rowcolors{1}{gray!10}{white}  

\begin{tabularx}{\linewidth}{|>{\centering\arraybackslash}p{0.06\textwidth}|>{\centering\arraybackslash}p{0.13\textwidth}|>{\centering\arraybackslash}p{0.13\textwidth}|>{\centering\arraybackslash}p{0.13\textwidth}|>{\centering\arraybackslash}p{0.13\textwidth}|>{\centering\arraybackslash}p{0.13\textwidth}|>{\centering\arraybackslash}p{0.13\textwidth}|}
\hline
\rowcolor{gray!20}  
\textbf{Scenario ID}& 
\textbf{RMSE X \newline (Kinematic)} & \textbf{RMSE X \newline (Map Matching)} & \textbf{Percentage \newline of Improvement X} & 
\textbf{RMSE Y \newline (Kinematic)} & \textbf{RMSE Y \newline (Map Matching)} & \textbf{Percentage \newline of Improvement Y} \\ \hline 
1 & \cellcolor{blue!15}0.8859 & \cellcolor{green!15}0.7044 & \cellcolor{white}20.49   & \cellcolor{blue!15}0.3846 & \cellcolor{green!15}0.9138 & \cellcolor{red!20}-137.60   \\ \hline 
2 & \cellcolor{blue!15}3.6304 & \cellcolor{green!15}0.3522 & \cellcolor{white}90.30   & \cellcolor{blue!15}3.8072 & \cellcolor{green!15}0.8588 & \cellcolor{white}77.44   \\ \hline 
3 & \cellcolor{blue!15}0.5296 & \cellcolor{green!15}0.7897 & \cellcolor{red!20}-49.11  & \cellcolor{blue!15}0.3183 & \cellcolor{green!15}0.4321 & \cellcolor{red!20}-35.75   \\ \hline 
4 & \cellcolor{blue!15}11.6885 & \cellcolor{green!15}0.7722 & \cellcolor{white}93.39   & \cellcolor{blue!15}13.3560 & \cellcolor{green!15}0.3456 & \cellcolor{white}97.41   \\ \hline 
5 & \cellcolor{blue!15}3.7951 & \cellcolor{green!15}0.2119 & \cellcolor{white}94.42   & \cellcolor{blue!15}15.8010 & \cellcolor{green!15}0.5814 & \cellcolor{white}96.32   \\ \hline 
6 & \cellcolor{blue!15}24.2994 & \cellcolor{green!15}0.3772 & \cellcolor{white}98.45   & \cellcolor{blue!15}7.9057 & \cellcolor{green!15}0.2785 & \cellcolor{white}96.48   \\ \hline 
\end{tabularx}

\renewcommand{\arraystretch}{1}  
\end{table*}

\begin{figure}
    \centering
    \includegraphics[width = \linewidth]{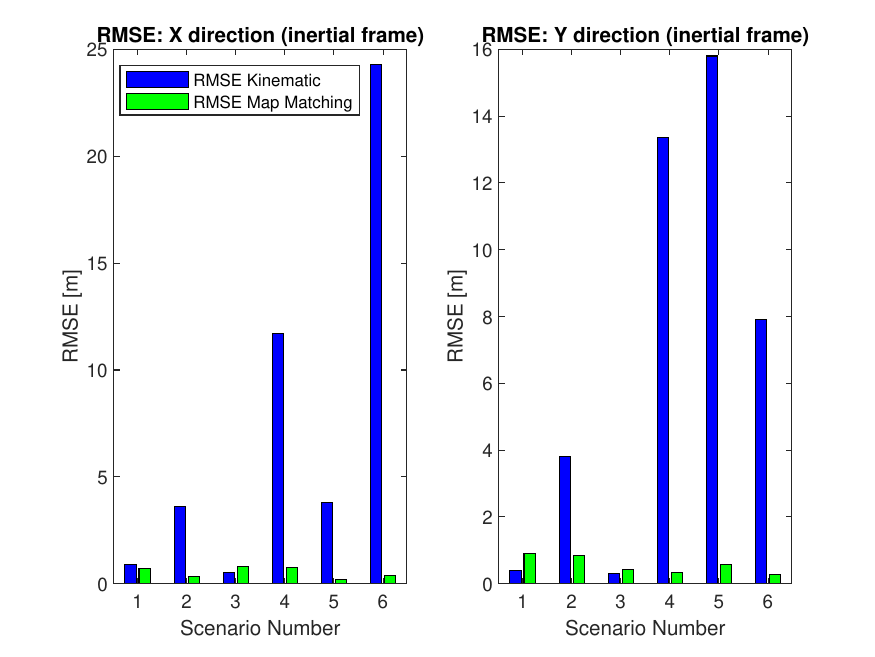}
    \caption{RMSE with respect to inertial frame}
    \label{fig:RMSE}
\end{figure}

The proposed method has been evaluated on six urban scenarios, each presenting a unique combination of road geometries and vehicle maneuvers as shown in Figure \ref{fig:Intersection_with_Map_Matching}.
The performance was assessed using the root mean square error (RMSE) in both X and Y directions, and the results were compared. It was noticed in Figure \ref{fig:Dead_Reckoning_Error_Accumulation} that the dead reckoning estimate had a significant drift error. By implementing the proposed method and parameter tuning, it is observed in Figure \ref{fig:Intersection_with_Map_Matching} that drift in the dead reckoning localization estimate is corrected in a spatial sense, and RMSE with respect to inertial frame of reference is improved (Figure \ref{fig:RMSE}) for all scenarios. 

Six distinct scenarios (road geometries and maneuvers) were tested. Figure \ref{fig:RMSE} illustrates the six scenarios and the spatially corrected map matching. Table \ref{tbl:rmse_comparison} shows the root mean square (RMSE) and the percentage of improvement in X and Y directions in all the scenarios, including temporal corrections. 

Scenarios 1 and 2 are conducted on a four-way intersection, where the vehicle takes a turn.
In Scenario 1, the test vehicle begins at a speed of 25 mph and enters a four-way intersection. Upon entering the intersection, the vehicle performs a sharp 90-degree right turn maneuver after that, it moves at 45 mph. The proposed method demonstrates a significant improvement of  90.30\% in the X direction (RMSE X: DR  3.6304 m, Map  0.3522 m). However, in the Y direction, the performance declined by -137.60\% (RMSE Y: DR 0.3846 m, Map  0.9138 m). In Scenario 2, the vehicle performs a left turn at the same four-way intersection. The proposed method yields improvements of 90.30\% in the X direction and 77.44\% in the Y direction.
In Scenario 3, the vehicle continues driving straight through the four-way intersection at a speed of 45 mph without performing any turns. The proposed method performance decline showed a -49.11\% change in the X direction (RMSE X: DR  0.5296 m, Map  0.7722 m) and a -35.75\% change in the Y direction (RMSE Y: DR  0.3183 m, Map  0.4321 m). In scenario 4, the vehicle drives at an average speed of 35 mph on a curved road and then enters a Y-intersection. The proposed method shows substantial improvements of 93.39\% and 97.41\% in X and Y directions, respectively. In scenario 5, the vehicle enters a roundabout and then takes the exit and goes straight. In this scenario  RMSE improved notably by 94.42\% in the X direction and 96.32\% in the Y direction. In scenario 6, the test vehicle enters a slip lane at 45 mph. The proposed method achieved improvement of 98.45\% in the X direction and 96.48\% in the Y direction.

The proposed map matching method reliably follows the ground truth GPS position in both spatial sense and temporal sense across all scenarios. This reliability underscores the efficiency of the arc length matching approach. Moreover, based on RMSE metrics, the proposed method performs well in complex road geometry scenarios and under varying speed conditions.
\section{Conclusion}
This paper introduces a novel iterative map matching approach for reliable vehicle localization in GPS-denied situations. The proposed method relies on the availability of spatial 2D map information of the geometry of the lane on which the vehicle is traveling.The results demonstrate good localization correction performance in both spatial and temporal sense, with RMSE improvement ranging from 20\% to 98\% across six distinct scenarios tested in this study as long as the correct lane on which the vehicle is traveling is known. The proposed method is effective at eliminating localization drift in GPS-denied environments. Future work will be focused on fusion with other additional sensors to enhance accuracy and support lane change.

\bibliographystyle{ieeetr}
\bibliography{arxiv_source/references_src}

\begin{thebibliography}{10}

\bibitem{vanet_survey}
F.~B. Günay, E.~Öztürk, T.~Çavdar, Y.~S. Hanay, and A.~u.~R. Khan, ``{Vehicular Ad Hoc Network (VANET) Localization Techniques: A Survey},'' {\em Archives of Computational Methods in Engineering}, 2021.

\bibitem{zhi}
Z.~Zheng, X.~Li, D.~Kong, J.~Hu, and Y.~Hu, ``{An effective fusion positioning methodology for land vehicles in GPS-denied environments using low-cost sensors},'' {\em Measurement Science and Technology}, vol.~34, p.~125102, aug 2023.

\bibitem{faroog}
F.~Ibrahim and K.~Nobukawa, ``{Positioning quality filter for the {V2X} technologies},'' Mar.~17 2020.
\newblock US Patent 10,591,608.

\bibitem{bajaj02}
R.~Bajaj, S.~Ranaweera, and D.~Agrawal, ``{{GPS}: location-tracking technology},'' {\em Computer}, vol.~35, no.~4, pp.~92--94, 2002.

\bibitem{alami}
A.~J. Alami, K.~El-Sayed, A.~Al-Horr, H.~Artail, and J.~Guo, ``{Improving the Car GPS accuracy using V2V and V2I Communications},'' in {\em {2018 IEEE International Multidisciplinary Conference on Engineering Technology (IMCET)}}, pp.~1--6, 2018.

\bibitem{big_survey}
S.~Kuutti, S.~Fallah, K.~Katsaros, M.~Dianati, F.~Mccullough, and A.~Mouzakitis, ``A survey of the state-of-the-art localization techniques and their potentials for autonomous vehicle applications,'' {\em IEEE Internet of Things Journal}, vol.~5, no.~2, pp.~829--846, 2018.

\bibitem{Kamijo}
S.~Kamijo, Y.~Gu, and H.~L., ``Autonomous vehicle technologies :localization and mapping,'' {\em Society of Electronic Information and Communications Technology Fundamentals Review}, vol.~9, no.~2, pp.~131--141, 2015.

\bibitem{Ward}
E.~Ward and J.~Folkesson, ``Vehicle localization with low cost radar sensors,'' in {\em 2016 IEEE Intelligent Vehicles Symposium (IV)}, pp.~864--870, 2016.

\bibitem{Rohani}
M.~Rohani, D.~Gingras, and D.~Gruyer, ``A novel approach for improved vehicular positioning using cooperative map matching and dynamic base station dgps concept,'' {\em IEEE Transactions on Intelligent Transportation Systems}, vol.~17, no.~1, pp.~230--239, 2016.

\bibitem{wahab}
A.~A. Wahab, A.~Khattab, and Y.~A. Fahmy, ``{Two-way TOA with limited dead reckoning for GPS-free vehicle localization using single RSU},'' in {\em 2013 13th International Conference on ITS Telecommunications (ITST)}, pp.~244--249, 2013.

\bibitem{Castorena}
J.~Castorena and S.~Agarwal, ``Ground-edge-based lidar localization without a reflectivity calibration for autonomous driving,'' {\em IEEE Robotics and Automation Letters}, vol.~3, no.~1, pp.~344--351, 2018.

\bibitem{Wolcott}
R.~W. Wolcott and R.~M. Eustice, ``Visual localization within lidar maps for automated urban driving,'' in {\em 2014 IEEE/RSJ International Conference on Intelligent Robots and Systems}, pp.~176--183, 2014.

\bibitem{ou}
C.-H. Ou, B.-Y. Wu, and L.~Cai, ``{GPS-free vehicular localization system using roadside units with directional antennas},'' {\em {Journal of Communications and Networks}}, vol.~21, no.~1, pp.~12--24, 2019.

\bibitem{wang}
W.~Jiangfeng, G.~Feng, Y.~Fei, and S.~Shaoxuan, ``{Design of wireless positioning algorithm of intelligent vehicle based on VANET},'' in {\em 2009 IEEE Intelligent Vehicles Symposium}, pp.~1098--1102, 2009.

\bibitem{sameh}
S.~W. Tawadrous and H.~A. Troemel~Jr, ``{Using ranging over {C-V2X} to supplement and enhance {GPS} performance},'' Feb.~25 2020.
\newblock US Patent 10,575,151.

\bibitem{ahmed}
A.~N. Ahmed, S.~Eckelmann, A.~Anwar, T.~Trautmann, and P.~Hellinckx, ``{"Lane Marking Detection Using LiDAR Sensor"},'' in {\em Advances on P2P, Parallel, Grid, Cloud and Internet Computing} (L.~Barolli, M.~Takizawa, T.~Yoshihisa, F.~Amato, and M.~Ikeda, eds.), (Cham), pp.~301--310, Springer International Publishing, 2021.

\bibitem{ohta}
H.~Ohta, N.~Akai, E.~Takeuchi, S.~Kato, and M.~Edahiro, ``Pure pursuit revisited: Field testing of autonomous vehicles in urban areas,'' in {\em {2016 IEEE 4th International Conference on Cyber-Physical Systems, Networks, and Applications (CPSNA)}}, pp.~7--12, 2016.

\bibitem{welte}
A.~Welte, P.~Xu, and P.~Bonnifait, ``{Four-Wheeled Dead-Reckoning Model Calibration using RTS Smoothing},'' in {\em 2019 International Conference on Robotics and Automation (ICRA)}, pp.~312--318, 2019.

\bibitem{freydin}
M.~Freydin and B.~Or, ``{Learning Car Speed Using Inertial Sensors for Dead Reckoning Navigation},'' {\em IEEE Sensors Letters}, vol.~6, no.~9, pp.~1--4, 2022.

\bibitem{carlson}
C.~R. Carlson, J.~C. Gerdes, and J.~D. Powell, ``{Error Sources When Land Vehicle Dead Reckoning with Differential Wheelspeeds},'' {\em Navigation}, vol.~51, no.~1, pp.~13--27, 2004.

\bibitem{brossard}
M.~Brossard, A.~Barrau, and S.~Bonnabel, ``{{AI-IMU} Dead-Reckoning},'' {\em IEEE Transactions on Intelligent Vehicles}, vol.~5, no.~4, pp.~585--595, 2020.

\bibitem{kaise}
Y.~Kaise, ``{Reckoning system using self reckoning combined with radio reckoning},'' Apr.~14 1998.
\newblock US Patent 5,740,049.

\bibitem{cao}
S.~Cao, Y.~Jin, T.~Trautmann, and K.~Liu, ``{Design and Experiments of Autonomous Path Tracking Based on Dead Reckoning},'' {\em Applied Sciences}, vol.~13, no.~1, 2023.

\bibitem{Kalman1960}
R.~E. Kalman, ``{A New Approach to Linear Filtering and Prediction Problems},'' {\em Journal of Basic Engineering}, vol.~82, pp.~35--45, 03 1960.

\bibitem{Simon2006}
D.~Simon, {\em Frontmatter}.
\newblock John Wiley \& Sons, Ltd, 2006.

\bibitem{Julier1997}
S.~J. Julier and J.~K. Uhlmann, ``{New extension of the Kalman filter to nonlinear systems},'' in {\em Signal Processing, Sensor Fusion, and Target Recognition VI} (I.~Kadar, ed.), vol.~3068, pp.~182 -- 193, International Society for Optics and Photonics, SPIE, 1997.

\bibitem{Ristic2004}
B.~Ristic, S.~Arulampalam, and N.~J. Gordon, {\em Beyond the Kalman Filter: Particle Filters for Tracking Applications}.
\newblock Artech House, 2004.

\bibitem{novatel}
Hexagon, ``{PwrPak7D}™.''

\bibitem{modern_robotics}
K.~M. Lynch and F.~C. Park, {\em {Modern Robotics: Mechanics, Planning, and Control}}.
\newblock Cambridge University Press, 2017.

\bibitem{geocoordinates_book}
P.~A. Longley, M.~F. Goodchild, D.~J. Maguire, and D.~W. Rhind, {\em {Geographical information systems and science}}.
\newblock J. Wiley \& sons., 2005.

\bibitem{spivak}
M.~Spivak, {\em {A Comprehensive Introduction to Differential Geometry}}.
\newblock {Publish or Perish}, 3rd~ed., 1999.

\end{thebibliography}

\section{Biography Section}
\begin{IEEEbiography}[{\includegraphics[width=1in,height=1.25in,clip,keepaspectratio]{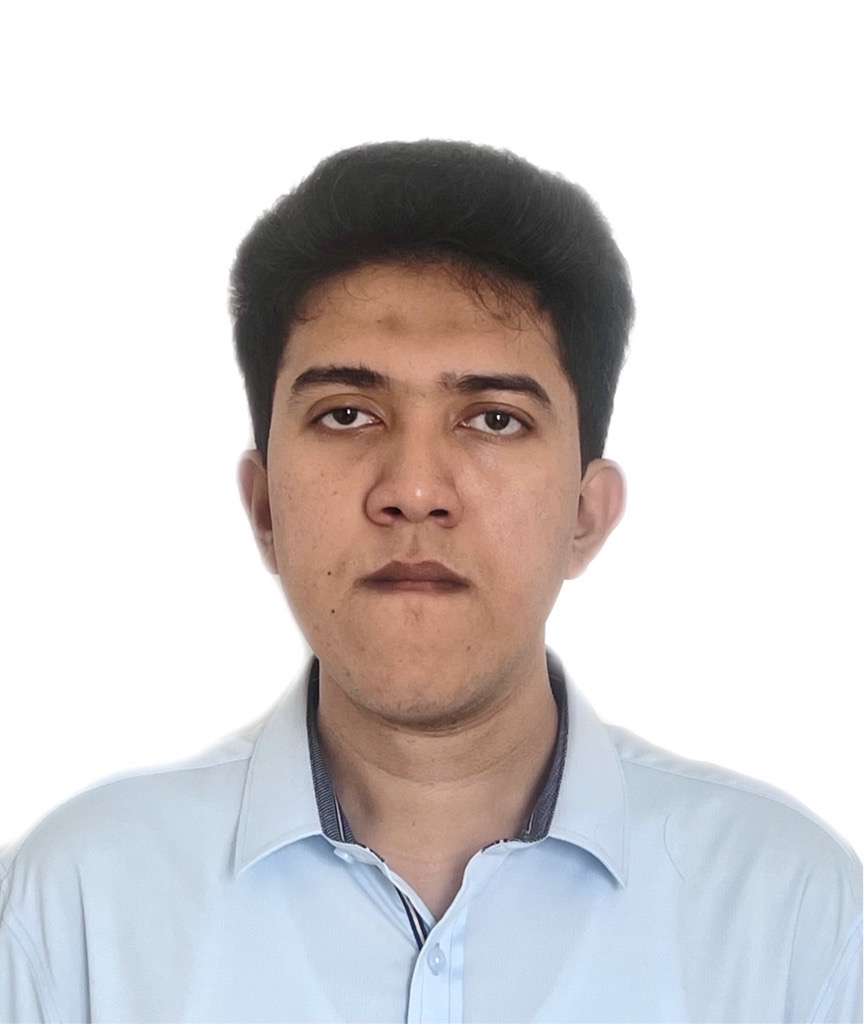}}]{Nur Uddin Javed}
received a B.S. degree in Electrical and Electronics Engineering from University Putra Malaysia (UPM) in 2022. He is currently pursuing an M.S. degree in Electrical and Computer Engineering at The Ohio State University and is a Graduate Research Associate at the Center for Automotive Research, The Ohio State University, Columbus, OH, USA. His research interests include automation systems, intelligent transportation systems, and robotics.
\end{IEEEbiography}

\begin{IEEEbiography}[{\includegraphics[width=1in,height=1.25in,clip,keepaspectratio]{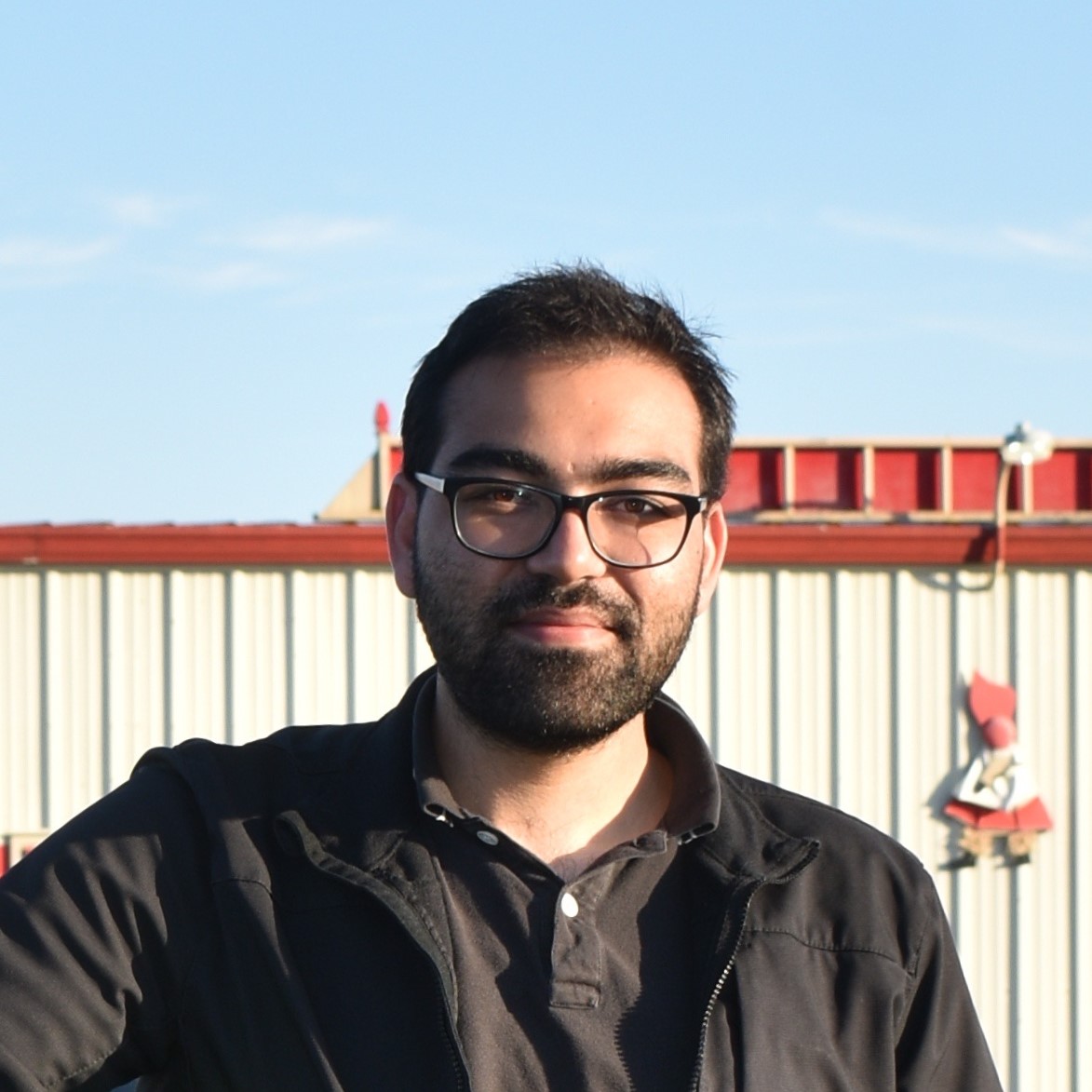}}]{Yuvraj Singh} (Student Member, IEEE) is a Mechanical Engineering Ph.D. student at the Ohio State University, Columbus OH, USA focusing on dynamics and control of automated vehicles. He received his B.E. degree in Mechanical engineering at Thapar Institute of Engineering and Technology, India in 2016. Then, he received his M.S. degree in mechanical engineering at The Ohio State University in 2020. He is currently a Graduate Research Associate at the Center for Automotive Research, Ohio State University. His research interests include nonlinear systems, vehicle dynamics, robotics, nonholonomic control, and signal processing.
\end{IEEEbiography}

\begin{IEEEbiography}[{\includegraphics[width=1in,height=1.25in,clip,keepaspectratio]{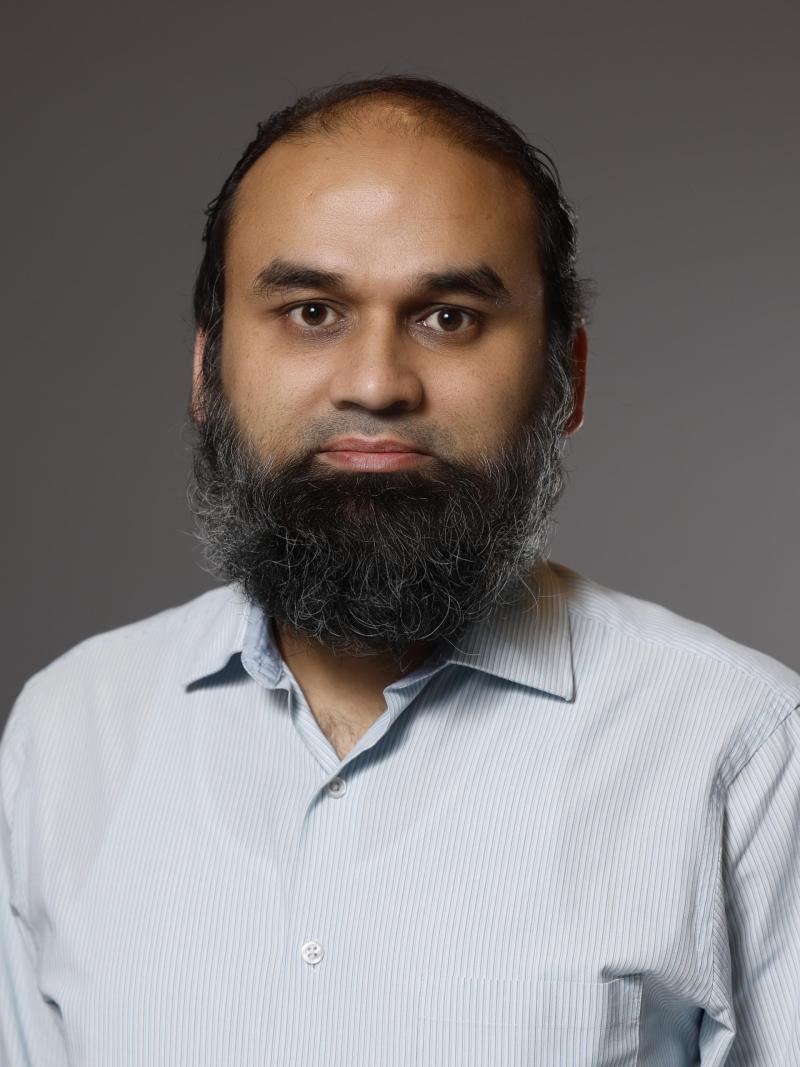}}]{Qadeer Ahmed}
(Senior Member, IEEE) received the Ph.D. degree in control systems from Mohammad Ali Jinnah University, Islamabad, Pakistan, in 2011. He is currently an Associate Professor with the Mechanical and Aerospace Engineering Department and the Center for Automotive Research, The Ohio State University, Columbus, OH, USA. His research interests include control and diagnostics of automotive systems with a focus on their efficiency, safety, and security.

\end{IEEEbiography}




\end{document}